\documentclass[lettersize,journal]{IEEEtran}
\usepackage[dvipdfmx]{graphicx}
\usepackage[dvipdfmx]{color}
\usepackage{amsmath,amsfonts}
\usepackage{algorithmic}
\usepackage{algorithm}
\usepackage{array}
\usepackage[caption=false,font=normalsize,labelfont=sf,textfont=sf]{subfig}
\usepackage{textcomp}
\usepackage{stfloats}
\usepackage{url}
\usepackage{verbatim}
\usepackage{graphicx}
\usepackage{cite}
\usepackage{bm}
\usepackage{here}
\usepackage{scalefnt}
\usepackage{mathtools}
\usepackage{amssymb}
\newcommand{\sBs}[3]{\mbox{ \!\!}^\mathrm{#1}\bm{#2}_\mathrm{#3}}

\newcommand{\arms}[0]{ARMS\ }
\newcommand{\armss}[0]{ARMS's\ }

\begin{document}

\title{A Mobile Quad-Arm Robot ARMS: Wheeled-Legged Tripedal Locomotion and Quad-Arm Loco-Manipulation}

\author{
	Hisayoshi~Muramatsu,~\IEEEmembership{Member,~IEEE,}
	Keigo~Kitagawa,
	Jun~Watanabe,
	Yuika~Yoshimoto,
	and Ryohei~Hisashiki
	\thanks{
		This work was supported by JKA and Nagamori Foundations, Japan. (Corresponding author: Hisayoshi Muramatsu).
		H.~Muramatsu, K.~Kitagawa, J.~Watanabe, Y.~Yoshimoto, and R.~Hisashiki are with Mechanical Engineering Program, Hiroshima University, Higashihiroshima 739-8527, Japan (e-mail: muramatsu@hiroshima-u.ac.jp).
	}
}

\markboth{}%
{Muramatsu \MakeLowercase{\textit{et al.}}: A Mobile Quad-Arm Robot ARMS: Wheeled-Legged Tripedal Locomotion and Quad-Arm Loco-Manipulation}


\maketitle
\begin{abstract}
This article proposes a mobile quad-arm robot: ARMS, which unifies wheeled-legged tripedal locomotion, wheeled locomotion, and quad-arm loco-manipulation.
\armss four arms have different mechanisms and are partially designed to be general-purpose arms for the hybrid locomotion and loco-manipulation.
One three-degree-of-freedom (DOF) arm has an active wheel, which is used for wheeled-legged tripedal walking and wheeled driving with passive wheels attached to the torso.
Two three-DOF general-purpose arms are series elastic and used for wheeled-legged tripedal walking, object grasping, and manipulation.
The upper two-DOF arm is used for manipulation only; its position and orientation are determined by coordinating all arms.
Each motor is controlled by an angle controller and trajectory modification with angle, angular velocity, angular acceleration, and torque constraints.
\arms was verified with seven experiments involving joint control, wheeled-legged locomotion, wheeled locomotion and grasping, slope locomotion, block terrain locomotion, carrying a bag, and outdoor locomotion.
\end{abstract}


\section{Introduction}
Robots are expected to acquire general locomotion and general task capabilities to perform various tasks in various environments.
Despite the demand, most existing robots are typically task-oriented, such as legged robots for uneven terrain \cite{Kau2019,Kim2021,Lee2021,Arm2019,Petr2021}, wheeled robots for even terrain \cite{Takaki2013,8884120}, and humanoid robots for manipulation tasks in living spaces \cite{Kaneko2019,Englsberger2014,Stasse2017}.
For enhancement of versatility, robots capable of hybrid locomotion and loco-manipulation have been developed.
%

For unification of legged locomotion and wheeled locomotion (hybrid locomotion), quadrupedal robots, such as RoboSimian~\cite{Hebert_RoboSimian}, CENTAURO~\cite{Kashiri2019}, ANYmal~\cite{Bjelonic_2019,Mederios2020}, Momaro~\cite{Schwarz2016}, BIT-NAZA II~\cite{CHEN2021367}, and a wheel-legged robot~\cite{Nagano2019}, are capable of both legged and wheeled locomotion.
RoboSimian~\cite{Hebert_RoboSimian} has two active wheels on its body and two passive wheels on its limbs, and the other robots have active wheels attached to the tips of their legs.
Additionally, there are wheeled-legged bipedal robots~\cite{ZHANG2022100027,Zhang2019,Klemm2019,Qiu2022}, wheeled-legged hexapedal robot~\cite{Orozco2021}, and wheeled robots whose wheels can transform into legs~\cite{Tadakuma2010,Chen2017,Zheng2019,Kim2020,Sun2023} for the hybrid locomotion.
For unification of legged locomotion and manipulation (loco-manipulation), quadrupedal robots, such as RoboSimian~\cite{Hebert_RoboSimian} and ALPHRED~\cite{Hooks2020}, have four general-purpose limbs.
The limbs can perform both quadrupedal walking and manipulation, such as grasping and valve turning, without additional limbs.
These general-purpose limbs are effective for reducing the number of actuators.
%

As mentioned above, the versatility of previous robots was enhanced by unifying the legged locomotion and manipulation or the legged locomotion and wheeled locomotion, while robots that unify all the legged locomotion, wheeled locomotion, and manipulation are rare.
Although RoboSimian~\cite{Hebert_RoboSimian} realizes the legged locomotion, wheeled locomotion, and manipulation, uniquely, the quadrupedal robot is redundant.
Accordingly, there is room for reducing the number of actuators for robots that are capable of hybrid locomotion and loco-manipulation.
%

%
\begin{figure}[t]
	\begin{center}
		\includegraphics[width=0.8\hsize]{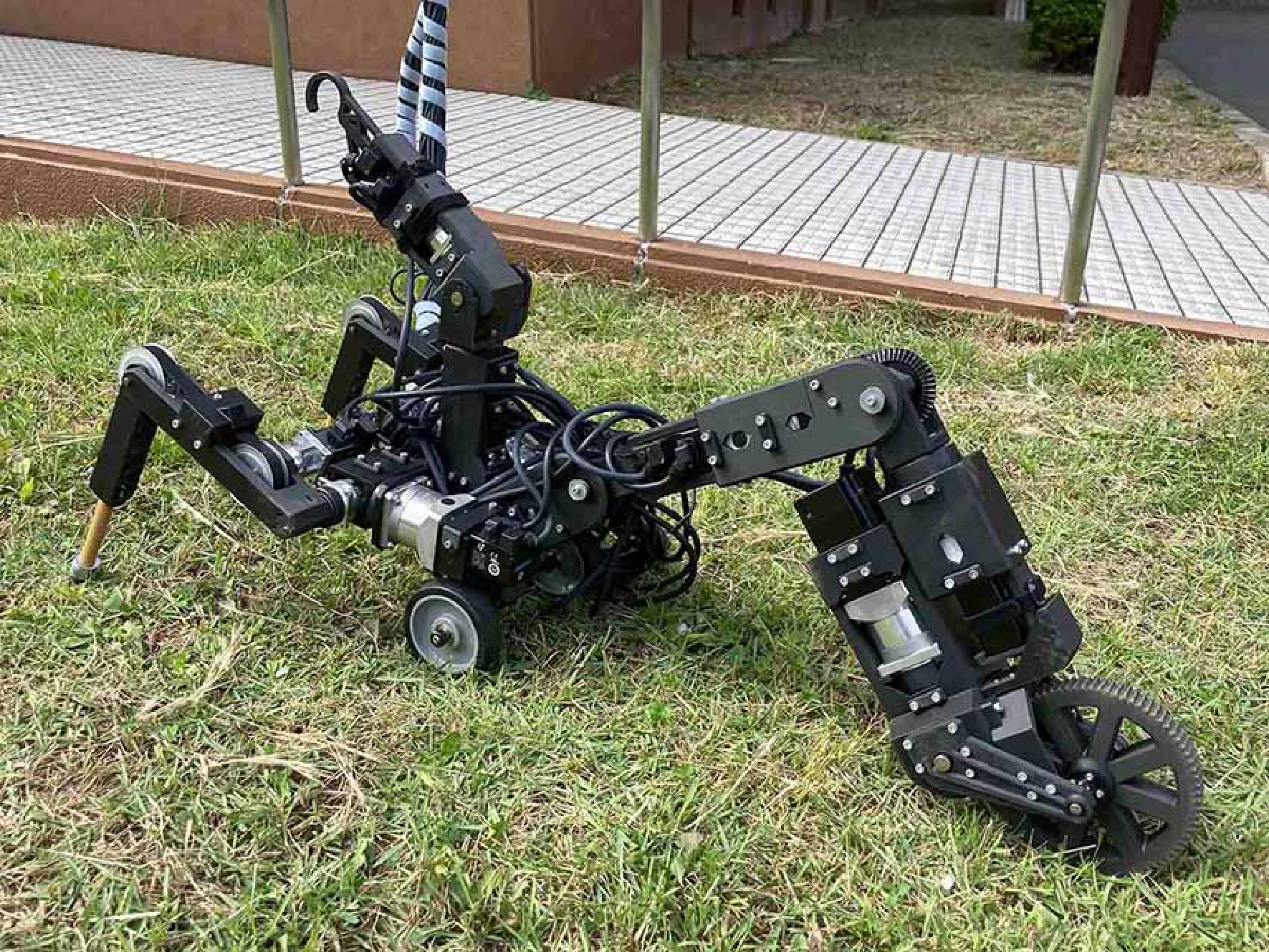}
	\end{center}
	\caption{Mobile quad-arm robot: ARMS.}\label{fig:ARMS}
\end{figure}
\begin{figure*}[t!]
	\footnotesize
	\begin{minipage}{0.5\hsize}
		\begin{center}
			\includegraphics[width=0.95\hsize]{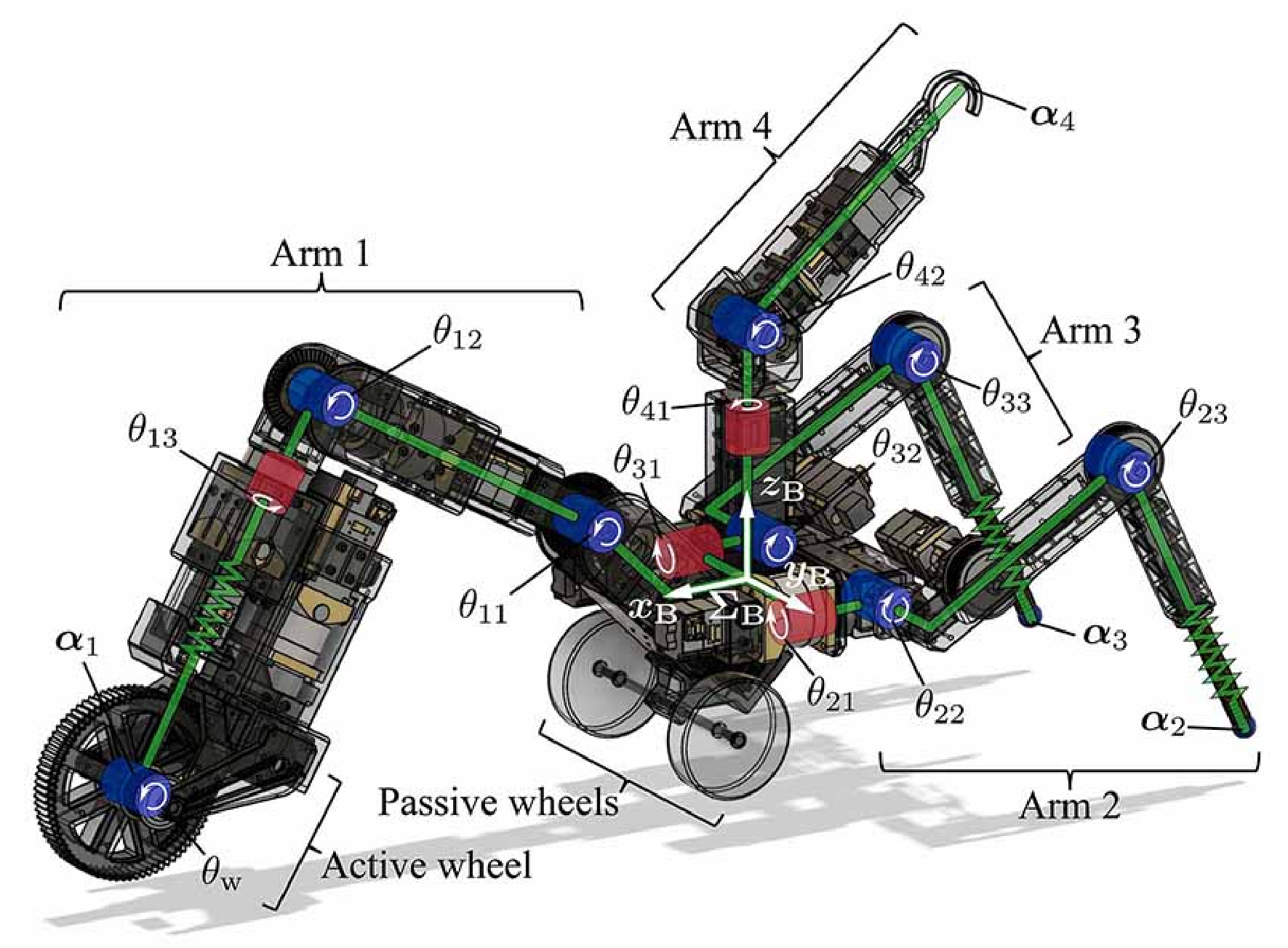}
		\end{center}
	\end{minipage}
	\begin{minipage}{0.5\hsize}
		\begin{center}
			\includegraphics[width=0.95\hsize]{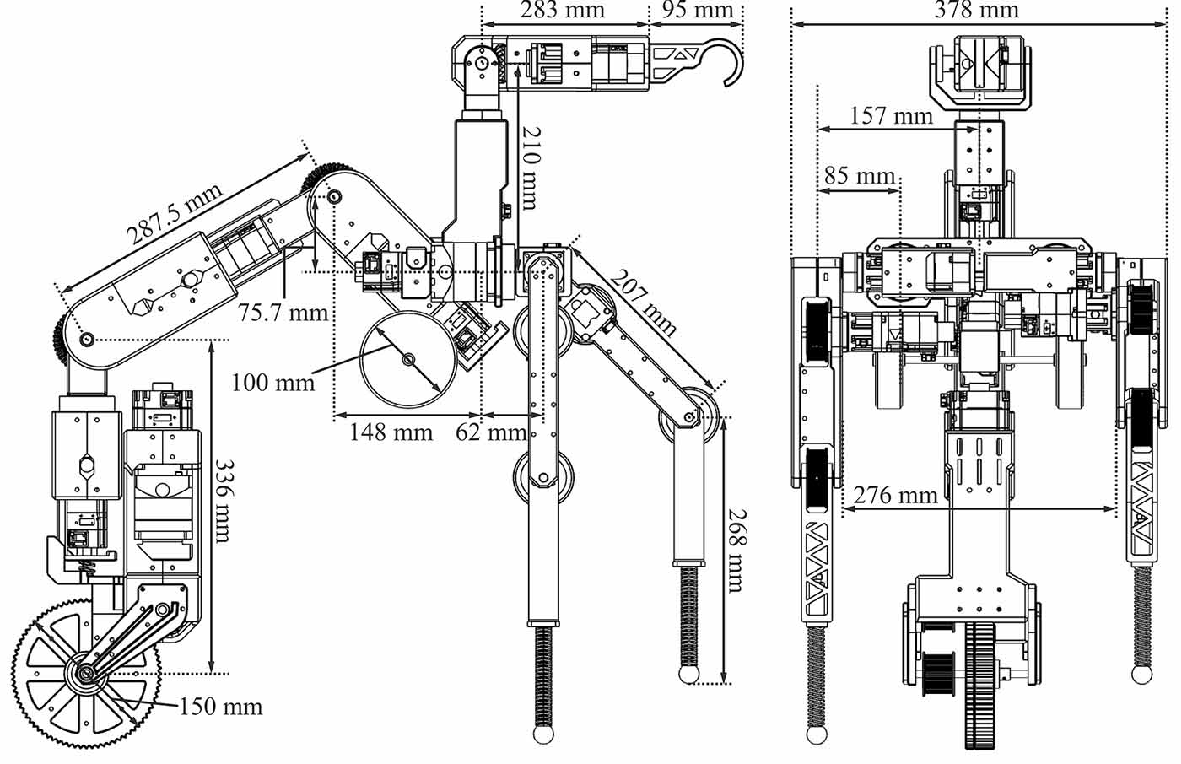}
		\end{center}
	\end{minipage}
	\begin{minipage}{0.5\hsize}
		\begin{center}
			(a)
		\end{center}
	\end{minipage}
	\begin{minipage}{0.5\hsize}
		\begin{center}
			(b)
		\end{center}
	\end{minipage}
	\begin{minipage}{0.33\hsize}
		\begin{center}
			\includegraphics[width=\hsize]{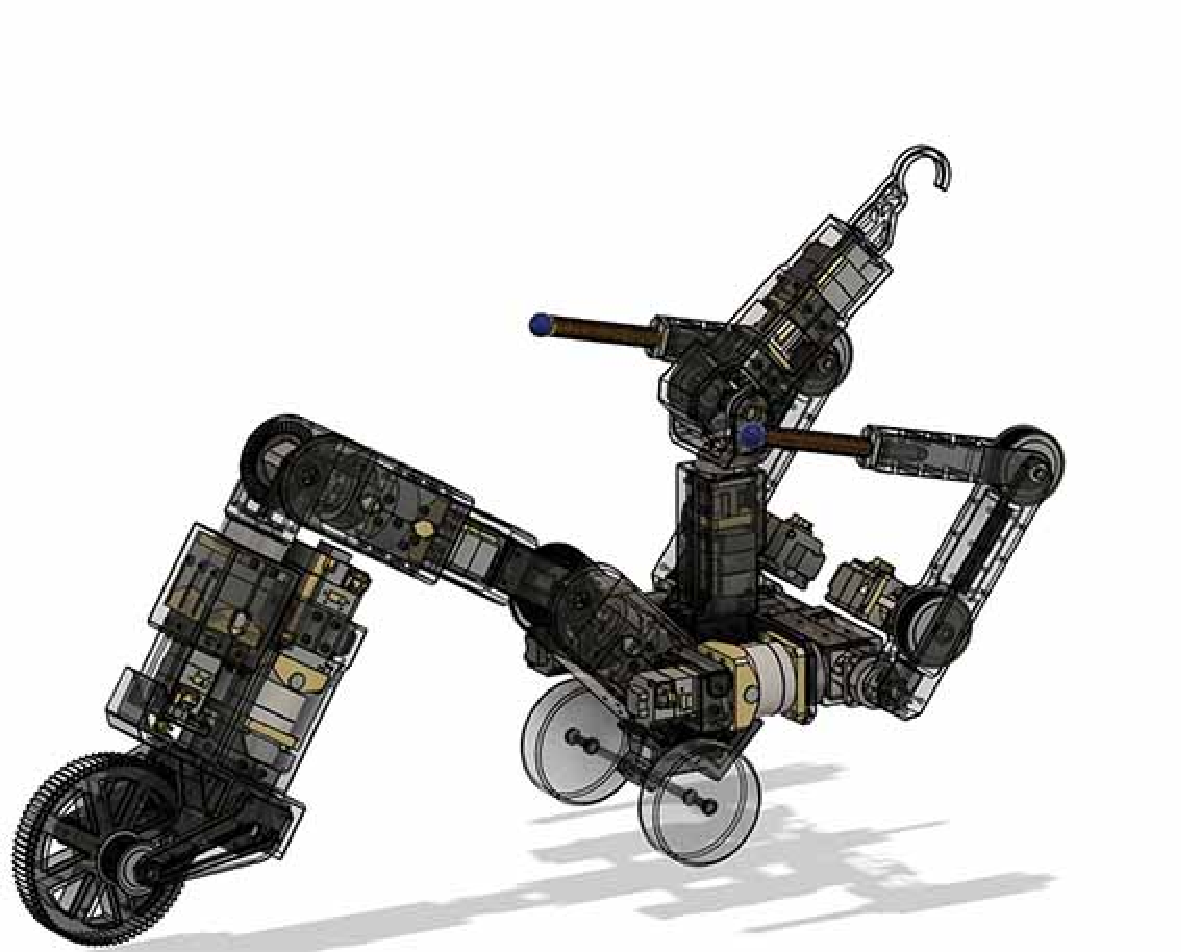}
		\end{center}
	\end{minipage}
	\begin{minipage}{0.33\hsize}
		\begin{center}
			\includegraphics[width=\hsize]{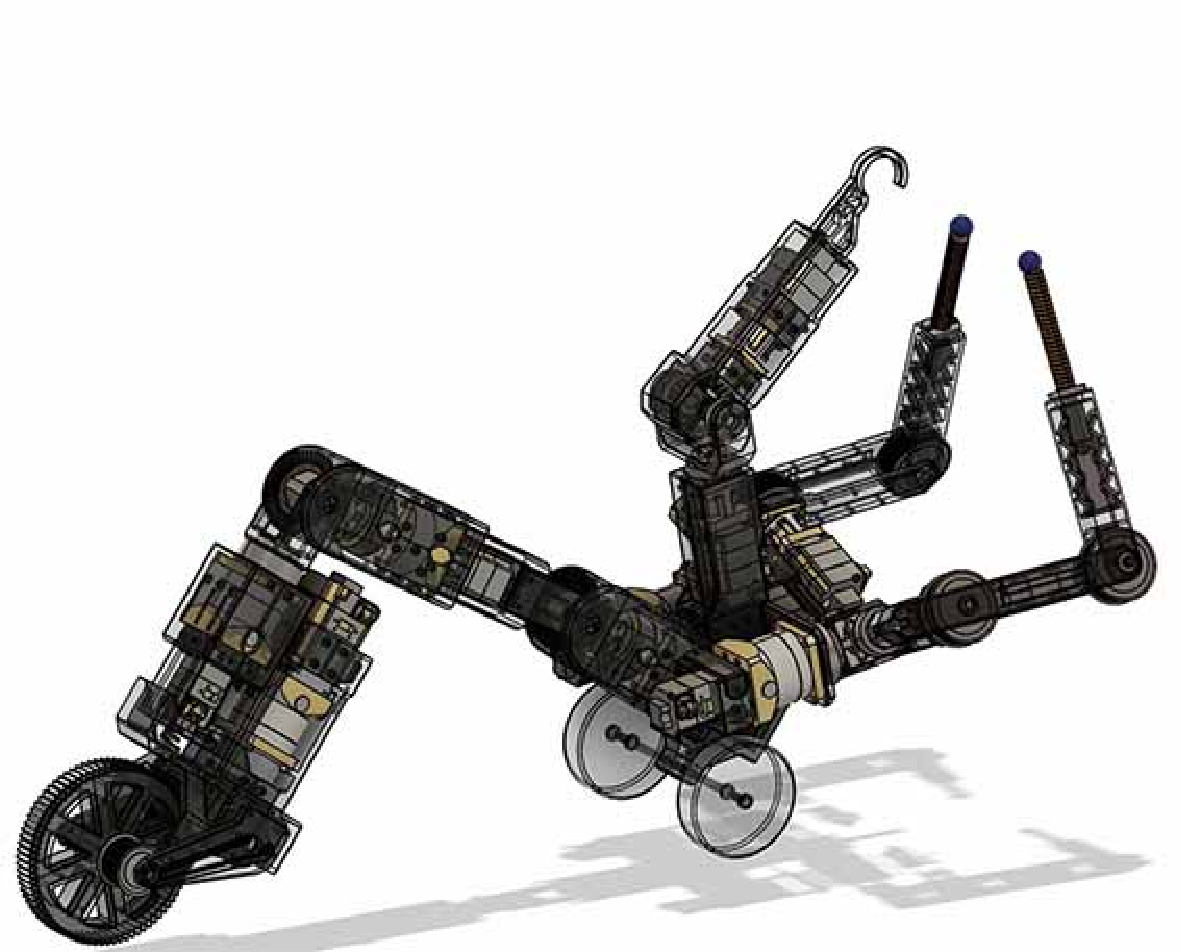}
		\end{center}
	\end{minipage}
	\begin{minipage}{0.33\hsize}
		\begin{center}
			\includegraphics[width=\hsize]{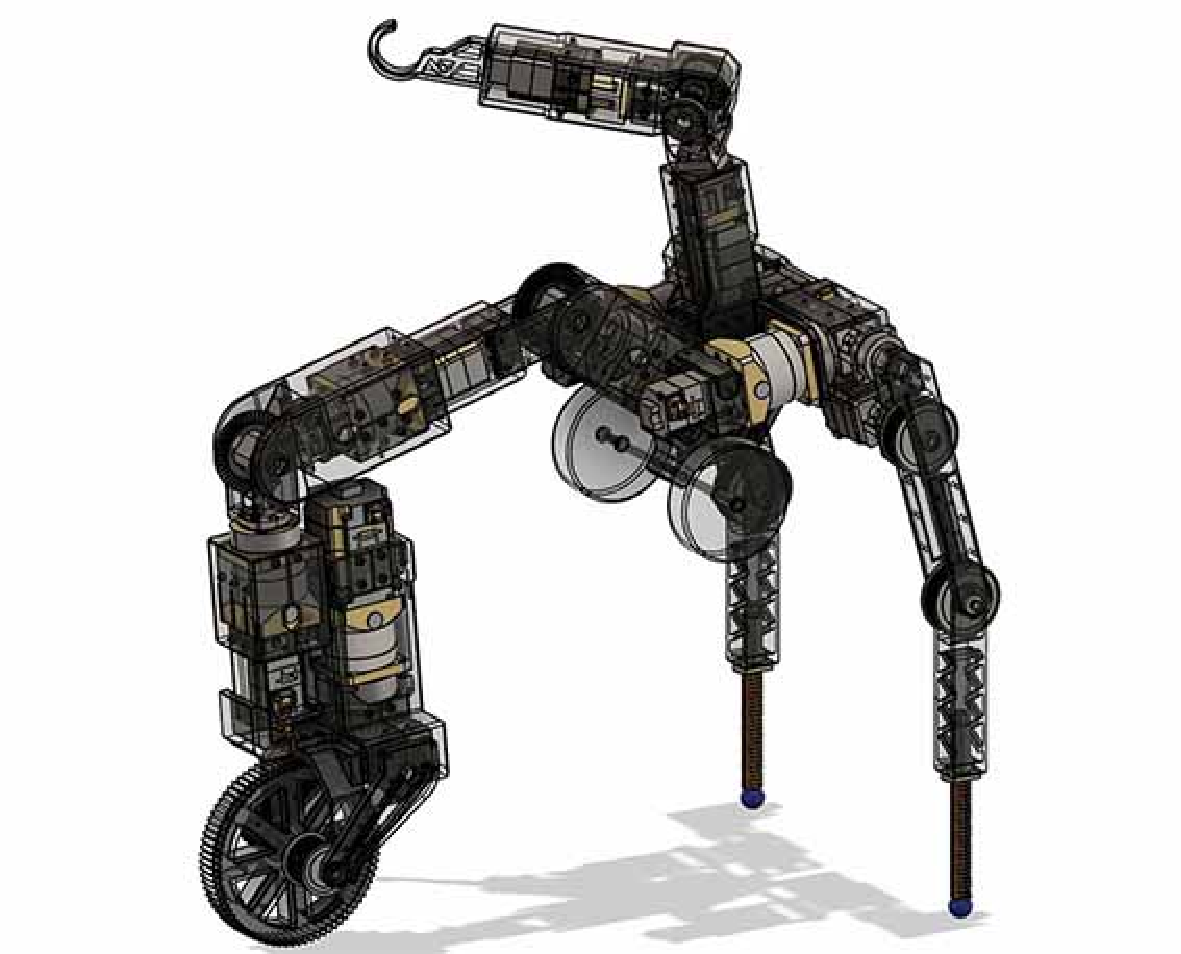}
		\end{center}
	\end{minipage}
	\begin{minipage}{0.33\hsize}
		\begin{center}
			(c)
		\end{center}
	\end{minipage}
	\begin{minipage}{0.33\hsize}
		\begin{center}
			(d)
		\end{center}
	\end{minipage}
	\begin{minipage}{0.33\hsize}
		\begin{center}
			(e)
		\end{center}
	\end{minipage}
	\caption{Configuration of ARMS with hybrid locomotion and loco-manipulation. (a) Wheeled-legged locomotion. (b) Dimensions in the side and back views. (c) Wheeled locomotion. (d) Wheeled locomotion and grasping. (e) Standing.}\label{fig:MW}
\end{figure*}
%

%
This article proposes a mobile quad-arm robot: ARMS that unifies wheeled-legged tripedal locomotion, wheeled locomotion, and quad-arm loco-manipulation.
For the hybrid locomotion and loco-manipulation, the four arms have different mechanisms, and two arms are designed to be general-purpose arms.
One three-degree-of-freedom (DOF) arm has an active wheel, which is used for wheeled-legged tripedal walking and wheeled driving with passive wheels attached to the torso.
Two three-DOF general-purpose arms are series elastic and are used for wheeled-legged tripedal walking, object grasping, and manipulation.
The upper two-DOF arm is used for manipulation only; its position and orientation are determined by coordinating all the arms.
\arms has 12 motors for the four arms and active wheel, and each motor is controlled by an angle controller and trajectory modification with angle, angular velocity, angular acceleration, and torque constraints.
Compared with the aforementioned robots, \arms achieves the hybrid locomotion and loco-manipulation, as the wheeled-legged tripedal locomotion, wheeled locomotion, and quad-arm loco-manipulation, with less DOF.
\armss tripedal legged locomotion using a single active wheel, passive wheels, and two legs is more stable than bipedal locomotion and less redundant than quadrupedal and hexapedal locomotions.
Additionally, \arms uses its arms for both locomotion and manipulation, which is similar to RoboSimian~\cite{Hebert_RoboSimian} and ALPHRED~\cite{Hooks2020} and different from CENTAURO~\cite{Kashiri2019}, ANYmal~\cite{Bjelonic_2019,Mederios2020}, Momaro~\cite{Schwarz2016}, BIT-NAZA II~\cite{CHEN2021367}, and the wheel-legged robot~\cite{Nagano2019}.
Furthermore, \arms has a manipulation-dedicated arm in contrast to RoboSimian~\cite{Hebert_RoboSimian}, and a single active wheel attached to the arm and two passive wheels at the torso in contrast to ALPHRED~\cite{Hooks2020}, CENTAURO~\cite{Kashiri2019}, ANYmal~\cite{Bjelonic_2019,Mederios2020}, Momaro~\cite{Schwarz2016}, BIT-NAZA II~\cite{CHEN2021367}, and the wheel-legged robot~\cite{Nagano2019}.

\section{Mechanical Design} \label{sec:2}
\arms is composed of a torso, four arms, a single active wheel, and two passive wheels.
The configuration of \arms is depicted in Fig.~\ref{fig:MW}(a), which shows 11 joints: three joints $\theta_{1j}$ for Arm~1, three joints $\theta_{2j}$ for Arm~2, three joints $\theta_{3j}$ for Arm~3, and two joints $\theta_{4j}$ for Arm~4; an active wheel $\theta_\mathrm{w}$ attached to the end-effector of Arm~1; and two passive wheels attached to the torso.
For shock absorption, Arm~1 has a spring between the motors for $\theta_{13}$ and $\theta_\mathrm{w}$, and Arms~2 and 3 have a series elastic structure.
The positions of the end-effectors of Arms~1, 2, 3, and 4 are $\bm{\alpha}_1$, $\bm{\alpha}_2$, $\bm{\alpha}_3$, and $\bm{\alpha}_4$, respectively, and the coordinate $\Sigma_\mathrm{B}$ is set to \armss torso.
Because each arm only has two or three DOFs, coordination of the four arms is necessary for controlling both the position and orientation of Arm 4's end-effector.
Fig.~\ref{fig:MW}(b) shows \armss dimensions in the side and back views in CAD images.
\arms is designed on the basis of the concepts of the hybrid locomotion and loco-manipulation.
Accordingly, it can perform wheeled-legged walking (Fig.~\ref{fig:MW}(a)), wheeled driving (Fig.~\ref{fig:MW}(c)), wheeled driving and grasping (Fig.~\ref{fig:MW}(d)), and standing (Fig.~\ref{fig:MW}(e)).
In the configurations, Arm~1 is used for the wheeled-legged walking, wheeled driving, and standing as a leg; Arms~2 and 3 are used for the legged walking, manipulation (grasping), and standing.
Meanwhile, Arm~4 is used for manipulation only.
The wheeled-legged walking is \armss basic configuration, in which \arms can move forward with the active and passive wheels' rotation and the gait of Arms~2 and 3.
For faster locomotion on flat terrain, the wheeled driving can move only with the active and passive wheels while \arms can grasp and manipulate an object by coordinating Arms~2, 3, and 4.
For performing tasks at an elevated position, \arms can stand with Arms~1, 2, and 3 and can manipulate an object using Arm 4's end-effector.
AC servo motors and amplifiers from Yaskawa Electric Corporation are employed as the actuators.
The motors used for Arm~1 are SGM7A-01A6AHC01 for all joints; Arms~2 and 3 use SGM7A-01A6AHC01 for joints $\theta_{21}$ and $\theta_{31}$ and SGM7A-A5A6AHC01 for the others; Arm~4 uses SGM7A-A5A6AHC01 for all joints; and the active wheel is driven by SGM7A-04A6AH101.
The joint specifications are presented in TABLE~\ref{tb:table1}.
Twelve amplifiers are selected from the following three types: SGD7S-R70-F-00-A, SGD7S-R90-F-00-A, and SGD7S-2R8-F-00-A.
The amplifiers are used in the torque control mode, in which they control the torques of the motors according to the torque command provided by a PC through a D/A PCI board (PCI-3340, Interface Corporation).
Meanwhile, the amplifiers send pulses given by the encoders attached to the motors to a counter PCI board (PCI-6205C, Interface Corporation), which is also attached to the PC.
The control algorithm presented in Section~\ref{sec:3} computes the torque command using the angle feedback from the counter board on the PC.
\begin{table}[t!]
	\caption{Joint specifications.}
	\label{tb:table1}
	\centering
	\scalebox{0.9}{
	\begin{tabular}{ccccccc}
	\hline
		& & \hspace{-2em}Min. & Max. & Max. &Max.\\
		& & \hspace{-2em}joint & joint & abs. & abs. \\
		& & \hspace{-2em}angle [deg] & angle [deg] & vel. [rad/s] & torque [Nm]\\
	\hline \hline
	Arm~1 &$\theta_{11}$ & \hspace{-2em}-110 & 110 & 14.97 & 20.8\\
		&$\theta_{12}$ & \hspace{-2em}-110 & 95 & 14.97 & 20.8\\
		&$\theta_{13}$ & \hspace{-2em}-180 & 180 & 14.97 & 20.8\\
	\hline
	Arm~2 &$\theta_{21}$ & \hspace{-2em}-90 & 30 & 14.97 & 20.8\\
		&$\theta_{22}$ & \hspace{-2em}-110 & 110 & 14.97 & 10.6\\
		&$\theta_{23}$ & \hspace{-2em}-120 & 120 & 14.97 & 10.6\\
	\hline
	Arm~3 &$\theta_{31}$ & \hspace{-2em}-30 & 90 & 14.97 & 20.8\\
		&$\theta_{32}$ & \hspace{-2em}-110 & 110 & 14.97 & 10.6\\
		&$\theta_{33}$ & \hspace{-2em}-120 & 120 & 14.97 & 10.6\\
	\hline
	Arm~4 &$\theta_{41}$ & \hspace{-2em}-180 & 180 & 14.97 & 10.6\\
		&$\theta_{42}$ & \hspace{-2em}-115 & 115 & 14.97 & 10.6\\
	\hline
	Wheel & $\theta_\mathrm{w}$ & \hspace{-2em}N/A & N/A & 62.83 & 20.1\\
	\hline
	\end{tabular}
	}
\end{table}

\begin{figure}[t!]
	\begin{center}
		\includegraphics[width=0.85\hsize]{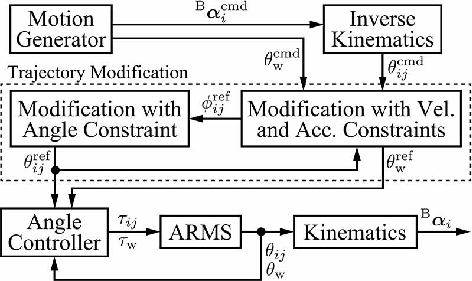}
	\end{center}
	\caption{Flow of the control system.}\label{fig:ctrl}
\end{figure}
\section{Control Design} \label{sec:3}
\arms is controlled as shown in Fig.~\ref{fig:ctrl}, which shows the motion generator, inverse kinematics, trajectory modification, and angle controller.
The motion generator provides the position command ${\sBs{B}{\alpha}{}}_{i}^{\mathrm{cmd}}$ for each arm and the rotation command $\theta_\mathrm{w}^\mathrm{cmd}$ for the active wheel, which is designed prior to experiments according to experimental scenarios.
The inverse kinematics gives the angle commands $\theta_{ij}^\mathrm{cmd}$ according to the position commands ${\sBs{B}{\alpha}{}}_{i}^{\mathrm{cmd}}$ for the arms.
Because the motors do not drive properly if its angle, velocity, acceleration, and torque constraints are not satisfied, the angle command needs to be modified not to exceed the limits.
First, the commands $\theta_{ij}^\mathrm{cmd}$ and $\theta_\mathrm{w}^\mathrm{cmd}$ are modified to the references $\phi_{ij}^\mathrm{ref}$ and $\theta_\mathrm{w}^\mathrm{ref}$, respectively, such that they are adapted for the velocity and acceleration constraints.
Subsequently, the reference $\phi_{ij}^\mathrm{ref}$ is modified to $\theta_{ij}^\mathrm{ref}$, which satisfies all the angle, velocity, and acceleration constraints.
To track the reference values, the motor angles $\theta_{ij}$ and $\theta_\mathrm{w}$ are controlled by proxy-based sliding mode control \cite{2006_Kikuuwe_PSMC,2010_Kikuuwe_PSMC,2014_Kikuuwe_AC} with consideration of their torque saturation, and the angle controllers output the torque $\tau_{ij}$ and $\tau_\mathrm{w}$ to the motors.
For simplicity, the subscripts $ij$ and $\mathrm{w}$ representing the joint are abbreviated below because the following computations are performed for each motor.
The trajectory modification involves two steps: modification with the velocity and acceleration constraints and modification with the angle constraint.
In the first step, the command $\theta_{k}^\mathrm{cmd}$ is modified to the reference $\phi_{k}^\mathrm{ref}$ as
\begin{align}
	\label{eq:mod}
	\phi_{k}^\mathrm{ref}&=\theta_{k-2}^\mathrm{cmd}+e_k.
\end{align}
The modification term $e_k$ is derived by quadratic programming that solves the following quadratic optimization problem with inequality constraints for the angular velocity and angular acceleration:
\begin{align}
	\label{eq:QP}
	&\hspace{-2em}\underset{e_k,\ e_{k+1},\ e_{k+2}}{\mathrm{minimize}}\ e_{k}^2+w_1T^2\dot{e}_{k+1}^2+w_2T^2\dot{e}_{k+2}^2\\
	\mathrm{s.t.}\ \
	&\phi_{k+h}=\theta_{k-2+h}^\mathrm{cmd}+e_{k+h}\notag\\
	&\phi_{k-1}=\theta_{k-1}^\mathrm{ref},\ \phi_{k-2}=\theta_{k-2}^\mathrm{ref}\notag\\
	&\dot{\phi}_{k+h}=(\phi_{k+h}-\phi_{k+h-1})/T\notag\\
	&\ddot{\phi}_{k+h}=(\phi_{k+h}-2\phi_{k+h-1}+\phi_{k+h-2})/T^2\notag\\
	&\dot{\theta}^\mathrm{min}\leq\dot{\phi}_{k+h}\leq\dot{\theta}^\mathrm{max},\
	\ddot{\theta}^\mathrm{min}\leq\ddot{\phi}_{k+h}\leq\ddot{\theta}^\mathrm{max}\notag\\
	&h=0,\ 1,\ 2.\notag
\end{align}
They are written in the discrete-time domain according to the backward difference with the sample $k$ and sampling time $T$, and the quadratic programming performs at each sampling.
The future modification terms $e_{k+1}$ and $e_{k+2}$ are considered to smooth the reference trajectory.
In the second step, the obtained reference angle $\phi_{k}^\mathrm{ref}$ is further modified according to the angle constraints without violating the velocity and acceleration constraints as follows
\begin{subequations}
	\label{eq:ddth_k_ref}
\begin{align}
	\label{eq:}
	\theta_k^\mathrm{ref}&=\left\{
		\begin{array}{cl}
			\phi_{k}^\mathrm{ref} & \hspace{-3em}\mathrm{if}\ {\theta}^\mathrm{min}\leq\hat{\phi}_{k+L}^\mathrm{ref}\leq{\theta}^\mathrm{max}\\
			\theta_{k-1}^\mathrm{ref}+T\dot{\theta}_{k-1}^\mathrm{ref}+T^2\ddot{\theta}_{k}^\mathrm{ref} & \mathrm{otherwise}
		\end{array}
	\right.\\
	\label{eq:ddth_k_ref_Sat}
	\ddot{\theta}_{k}^\mathrm{ref}&=\mathrm{sat}_{(\ddot{\theta}^\mathrm{min},\ \ddot{\theta}^\mathrm{max})}(-(\theta_{k-1}^\mathrm{ref}-\theta_{k-2}^\mathrm{ref})T^{-2}),
\end{align}
\end{subequations}
where
\begin{align*}
	&L =\mathrm{min}\{l\in \mathbb{Z}_{\geq0} | \dot{\phi}_{k}^\mathrm{ref}(\dot{\phi}_{k}^\mathrm{ref} + T\bar{\ddot{\theta}}_k + (l-1)T\underline{\ddot{\theta}}_k) \leq 0\}\\
	&\hat{\phi}_{k}^\mathrm{ref}=\phi_{k}^\mathrm{ref},\
	\hat{\phi}_{k+1}^\mathrm{ref}=\bm{C}(\bm{A}\Phi_{k}^\mathrm{ref}+\bm{B}\bar{\ddot{\theta}}_k)\\
	&\hat{\phi}_{k+L|L>1}^\mathrm{ref}=\bm{C}\bm{A}^L\Phi_{k}^\mathrm{ref}+\bm{C}\bm{A}^{L-1}\bm{B}\bar{\ddot{\theta}}_k+\bm{C}\sum_{i=0}^{L-2}\bm{A}^i\bm{B}\underline{\ddot{\theta}}_k\\
	&\Phi_{k}^\mathrm{ref}=\left[
		\begin{array}{c}
			\phi_{k}^\mathrm{ref}\\
			\dot{\phi}_{k}^\mathrm{ref}
		\end{array}
	\right],\
	\bm{A}=\left[
		\begin{array}{cc}
			1&T\\
			0&1
		\end{array}
	\right],\
	\bm{B}=\left[
		\begin{array}{c}
			0\\
			T
		\end{array}
	\right]\\
	&\bm{C}=\left[
		\begin{array}{cc}
			1&0
		\end{array}
	\right],\
	\mathrm{sat}_{(a,\ b)}(x)= \left\{
	\begin{array}{cl}
		a&\mathrm{if}\ x< a\\
		x&\mathrm{if}\ a\leq x\leq b\\
		b&\mathrm{if}\ x > b
	\end{array}
	\right.\\
	&\bar{\ddot{\theta}}_k=\left\{
	\begin{array}{cl}
		\ddot{\theta}^\mathrm{max}&\mathrm{if}\ \dot{\phi}_{k}^\mathrm{ref}>0\\
		0&\mathrm{if}\ \dot{\phi}_{k}^\mathrm{ref}=0\\
		\ddot{\theta}^\mathrm{min}&\mathrm{if}\ \dot{\phi}_{k}^\mathrm{ref}<0,
	\end{array}
	\right.\
	\underline{\ddot{\theta}}_k=\left\{
	\begin{array}{cl}
		\ddot{\theta}^\mathrm{min}&\mathrm{if}\ \dot{\phi}_{k}^\mathrm{ref}>0\\
		0&\mathrm{if}\ \dot{\phi}_{k}^\mathrm{ref}=0\\
		\ddot{\theta}^\mathrm{max}&\mathrm{if}\ \dot{\phi}_{k}^\mathrm{ref}<0.
	\end{array}
	\right.
\end{align*}
Here, $L$ denotes the deceleration time until sign reversal of the reference velocity $\phi_{k}^\mathrm{ref}$, which is composed of the initial maximum acceleration $\bar{\ddot{\theta}}_k$ at $k+1$ and subsequent maximum deceleration $\underline{\ddot{\theta}}_k$ from $k+2$ to $k+L$.
If the angle violates the angle constraints ${\theta}^\mathrm{min}\leq\hat{\phi}_{k+L}^\mathrm{ref}\leq{\theta}^\mathrm{max}$ after the deceleration, the acceleration $\ddot{\theta}_{k}^\mathrm{ref}$ is set to bring the velocity $\dot{\theta}_{k}^\mathrm{ref}$ to zero with acceleration saturation in accordance with \eqref{eq:ddth_k_ref_Sat}, which yields $\dot{\theta}_k^\mathrm{ref}=\dot{\theta}_{k-1}^\mathrm{ref}+T\ddot{\theta}_{k}^\mathrm{ref}=0$ during unsaturation and decelerates to zero during saturation.
The angle controller uses proxy-based sliding mode control \cite{2006_Kikuuwe_PSMC,2010_Kikuuwe_PSMC,2014_Kikuuwe_AC}, which seamlessly changes proportional-integral-derivative angle control with fast reference tracking, force saturation in the reaching mode, and smooth convergence to the reference angle after saturation in the sliding mode.
The continuous-time representation of the controller is
\begin{subequations}
	\label{eq:PSMC}
\begin{align}
	\label{eq:PSMC:pid}
	\tau&=M\ddot{p}-B\ddot{\alpha}-K\dot{\alpha}-L\alpha\\
	\label{eq:PSMC:smc}
	\tau&\in \mathrm{sgn}_{F}(-J\ddot{\beta}-H\dot{\beta}-\beta)\\
	\dot{\alpha}&= \theta-p,\
	\beta= p - \theta^\mathrm{ref},
\end{align}
\end{subequations}
where $M$ and $F$ denote the moment of inertia and force limit of the motors, respectively.
The proportional-integral-derivative controller \eqref{eq:PSMC:pid} is dominant during unsaturation, and the sliding mode controller \eqref{eq:PSMC:smc} based on the set-valued sign function
\begin{align*}
	\mathrm{sgn}_{F}(x)= & \left\{
	\begin{array}{cl}
		\{-F\}&\mathrm{if}\ x< 0\\
		{[-F,\ F]}&\mathrm{if}\ x=0\\
		\{F\}&\mathrm{if}\ x > 0
	\end{array}
	\right.
\end{align*}
is dominant during and after saturation.
By using the proxy-based sliding mode control, we can design an unsaturated response with the proportional-integral-derivative gains $B$, $K$, and $L$, can implement saturation as a part of the sliding mode controller, and can design the response after saturation with the gains $J$ and $H$ of the switching function $-J\ddot{\beta}-H\dot{\beta}-\beta$.
%

The control algorithm shown in Fig.~\ref{fig:ctrl} was implemented on a PC with the Linux OS and RTAI, which had two real-time threads: the thread for the angle controller with a sampling time of 0.25 ms and that for the other computations with a sampling time of 0.5 ms.

\begin{figure}[t!]
	\footnotesize
	\begin{center}
		\includegraphics[width=0.7\hsize]{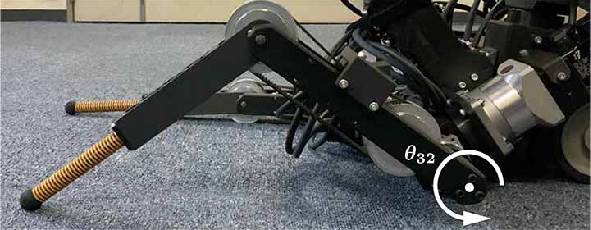}\\
		(a)\\
		\includegraphics[width=0.9\hsize]{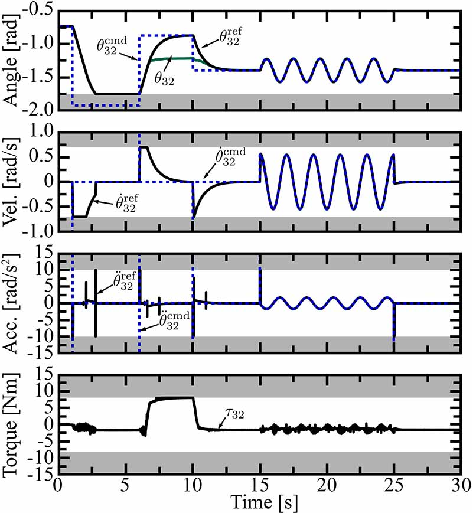}\\
		(b)
	\end{center}
	\caption{Exp. 1: Joint control with the trajectory modification \eqref{eq:mod}--\eqref{eq:ddth_k_ref} and proxy-based sliding mode controller \eqref{eq:PSMC}. (a) Contact motion from 6 to 10 s. (b) Results of the joint $\theta_{32}$. The bray bands represent upper and lower limits.}\label{fig:joint:exp}
\end{figure}
\begin{figure*}[t!]
	\begin{center}
		\includegraphics[width=0.925\hsize]{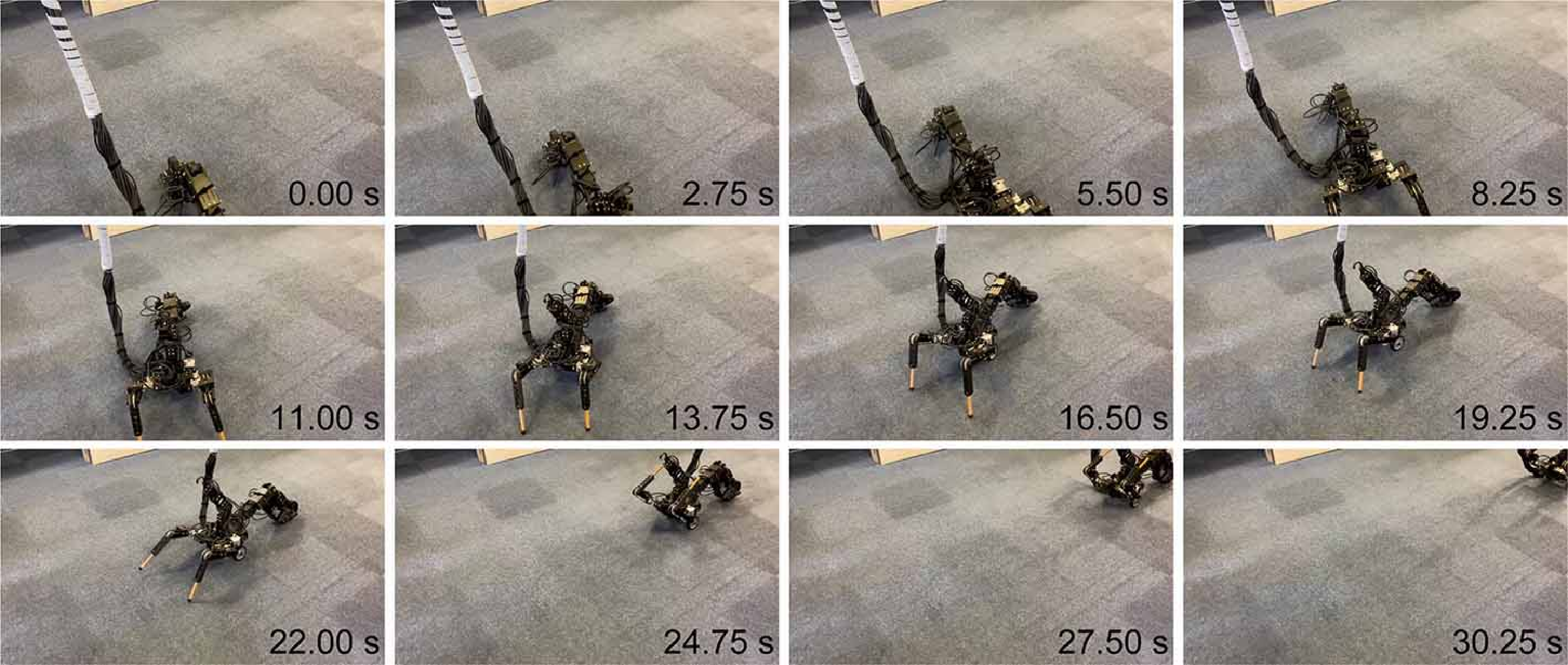}
	\end{center}
	\caption{Exp. 2: Wheeled-legged locomotion.}\label{fig:exp:WheeledLeggedLocomotion}
\end{figure*}
\begin{figure*}[t!]
	\begin{center}
		\includegraphics[width=0.925\hsize]{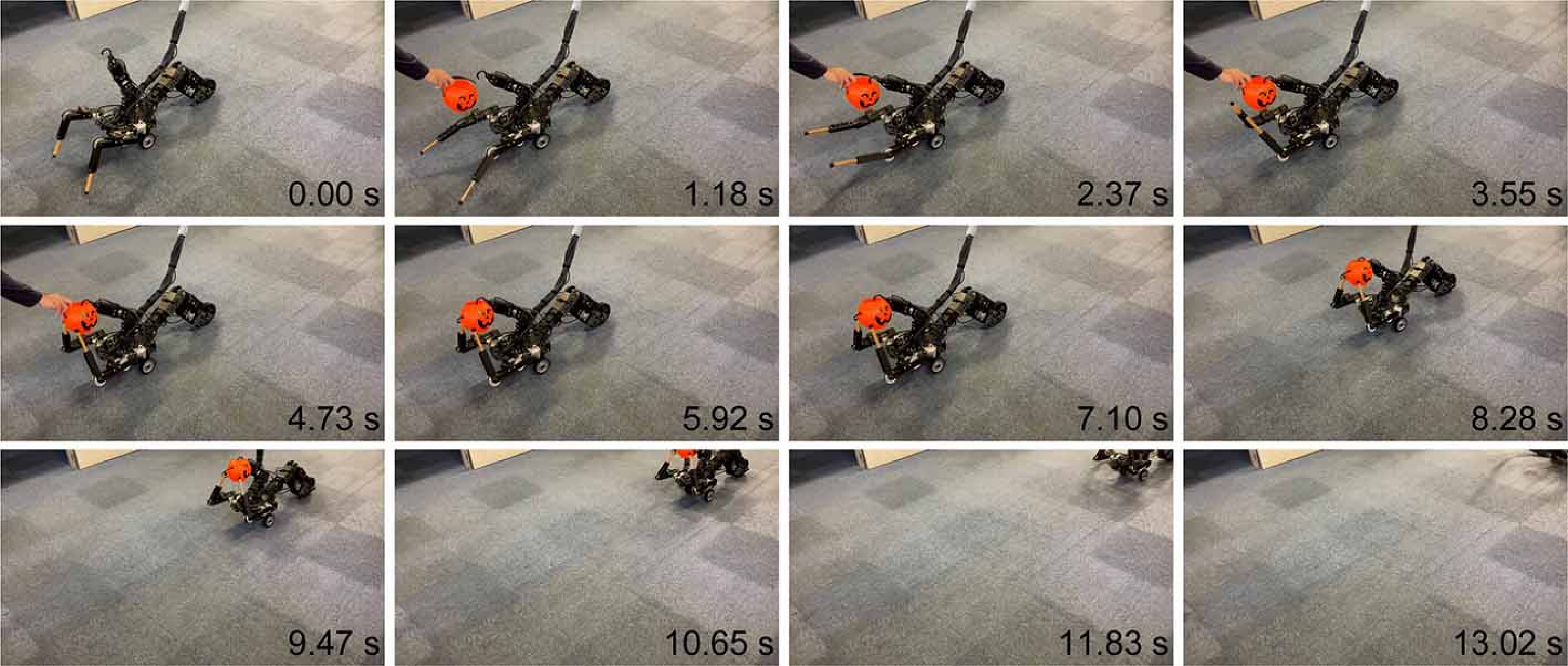}
	\end{center}
	\caption{Exp. 3: Wheeled locomotion and grasping.}\label{fig:exp:grasping}
\end{figure*}
\begin{figure*}[t!]
	\begin{center}
		\includegraphics[width=0.925\hsize]{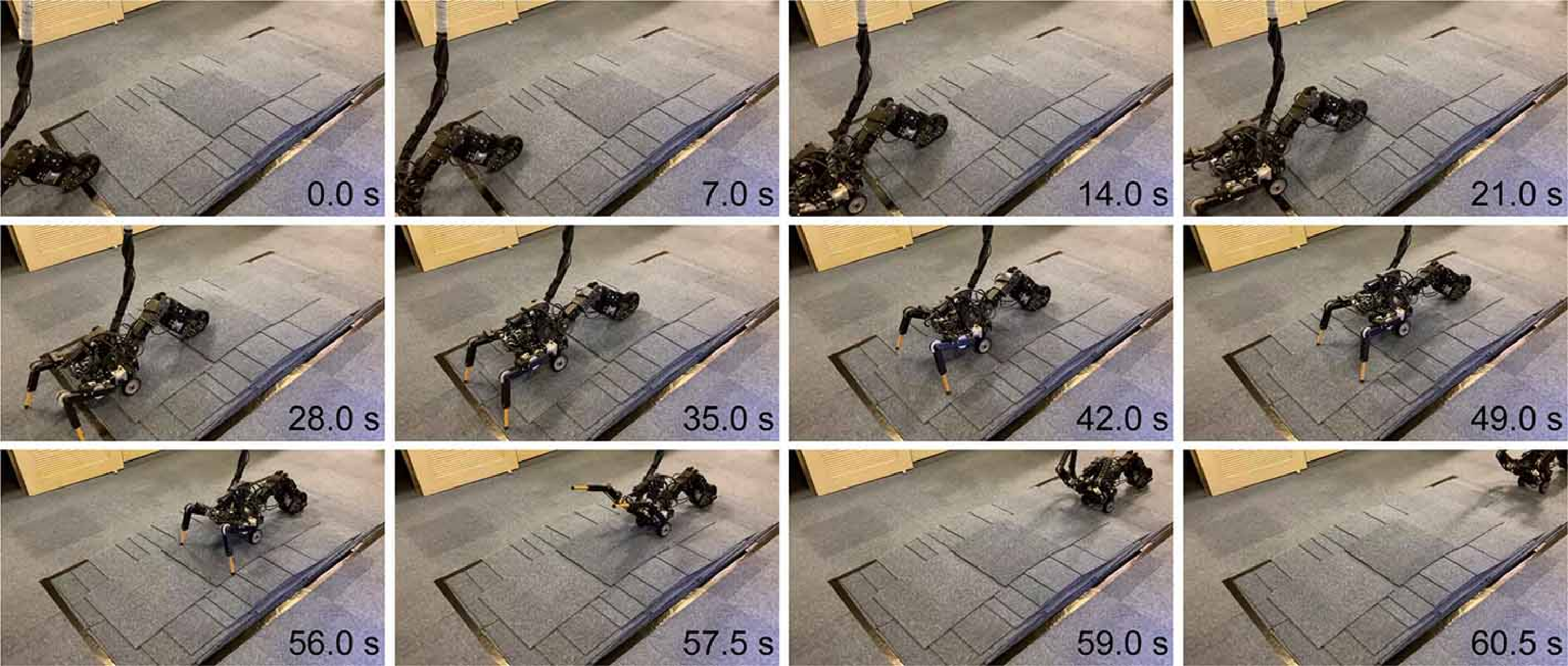}
	\end{center}
	\caption{Exp. 4: Slope locomotion.}\label{fig:exp:slope}
\end{figure*}
\begin{figure*}[t!]
	\begin{center}
		\includegraphics[width=0.925\hsize]{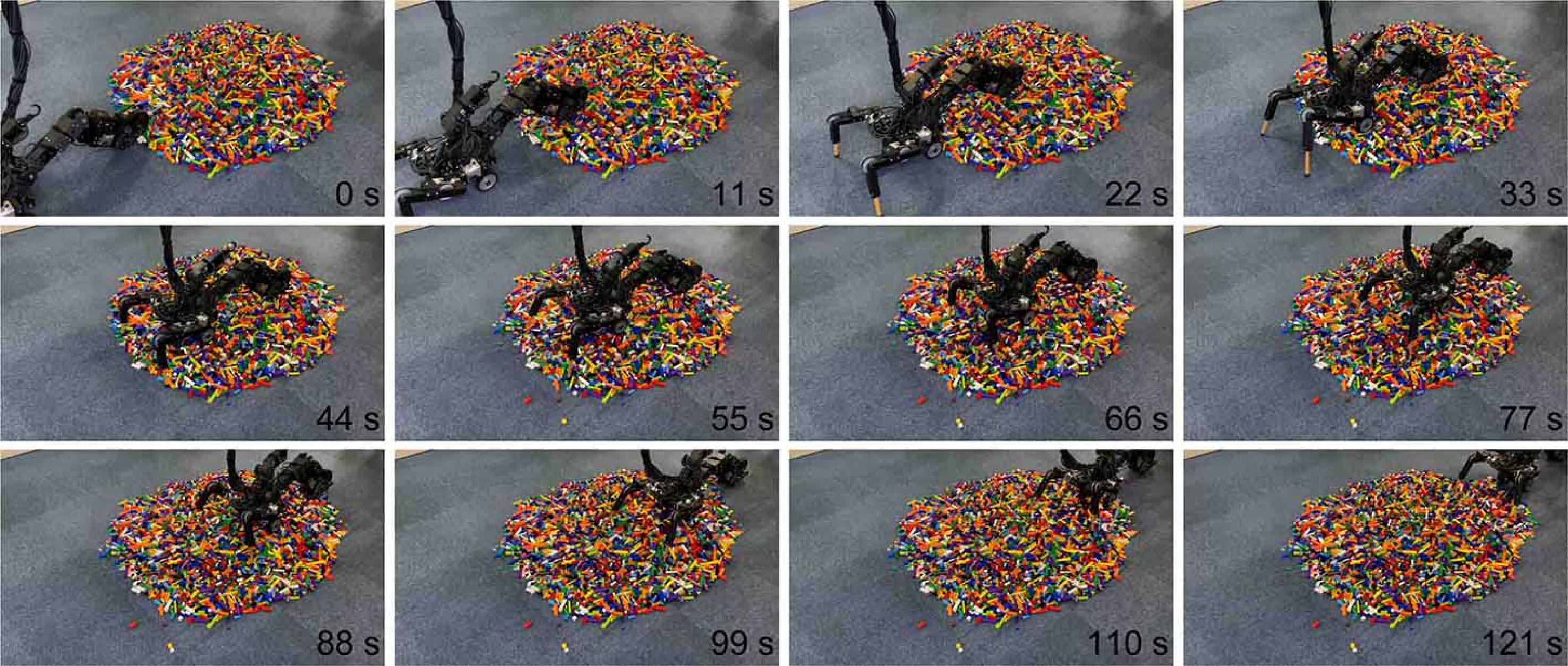}
	\end{center}
	\caption{Exp. 5: Block terrain locomotion.}\label{fig:exp:block}
\end{figure*}
\begin{figure*}[t!]
	\begin{center}
		\includegraphics[width=0.925\hsize]{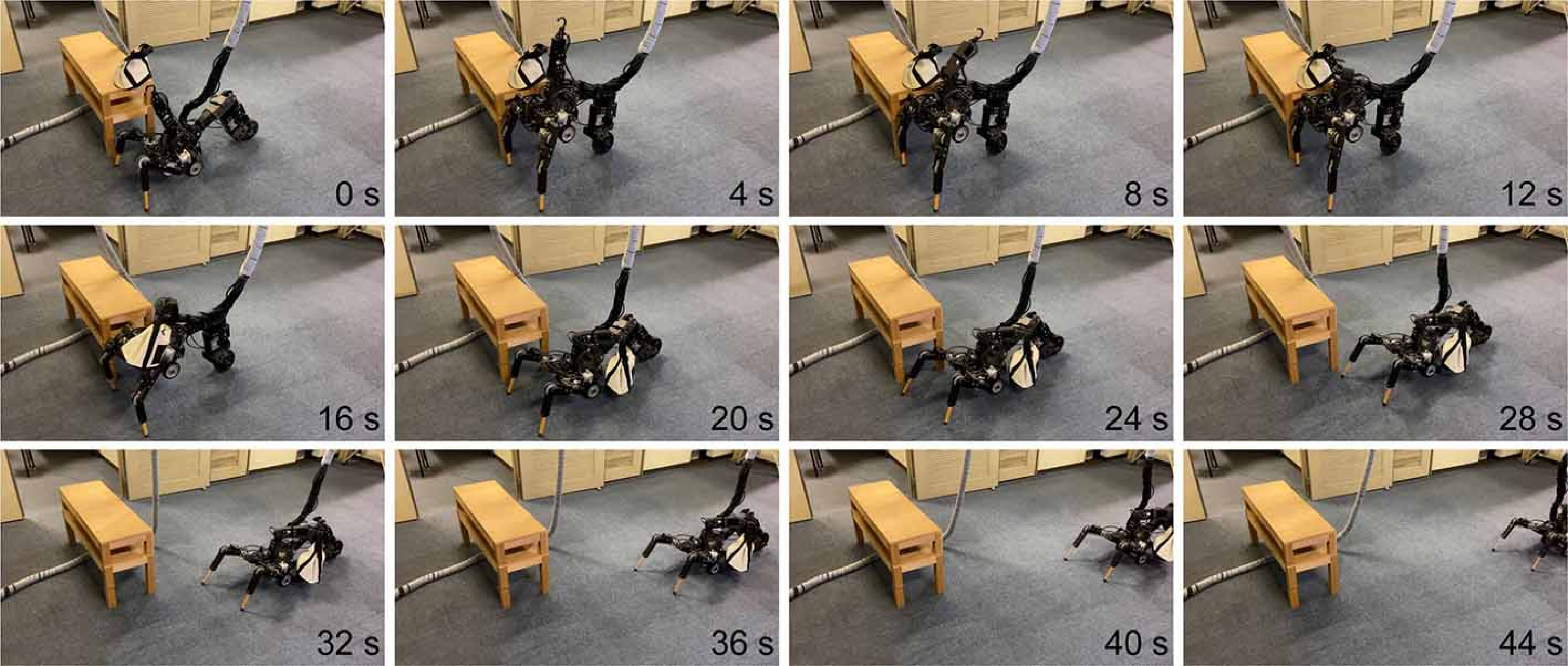}
	\end{center}
	\caption{Exp. 6: Carrying a bag.}\label{fig:exp:carry}
\end{figure*}
\begin{figure*}[t!]
	\begin{center}
		\includegraphics[width=0.925\hsize]{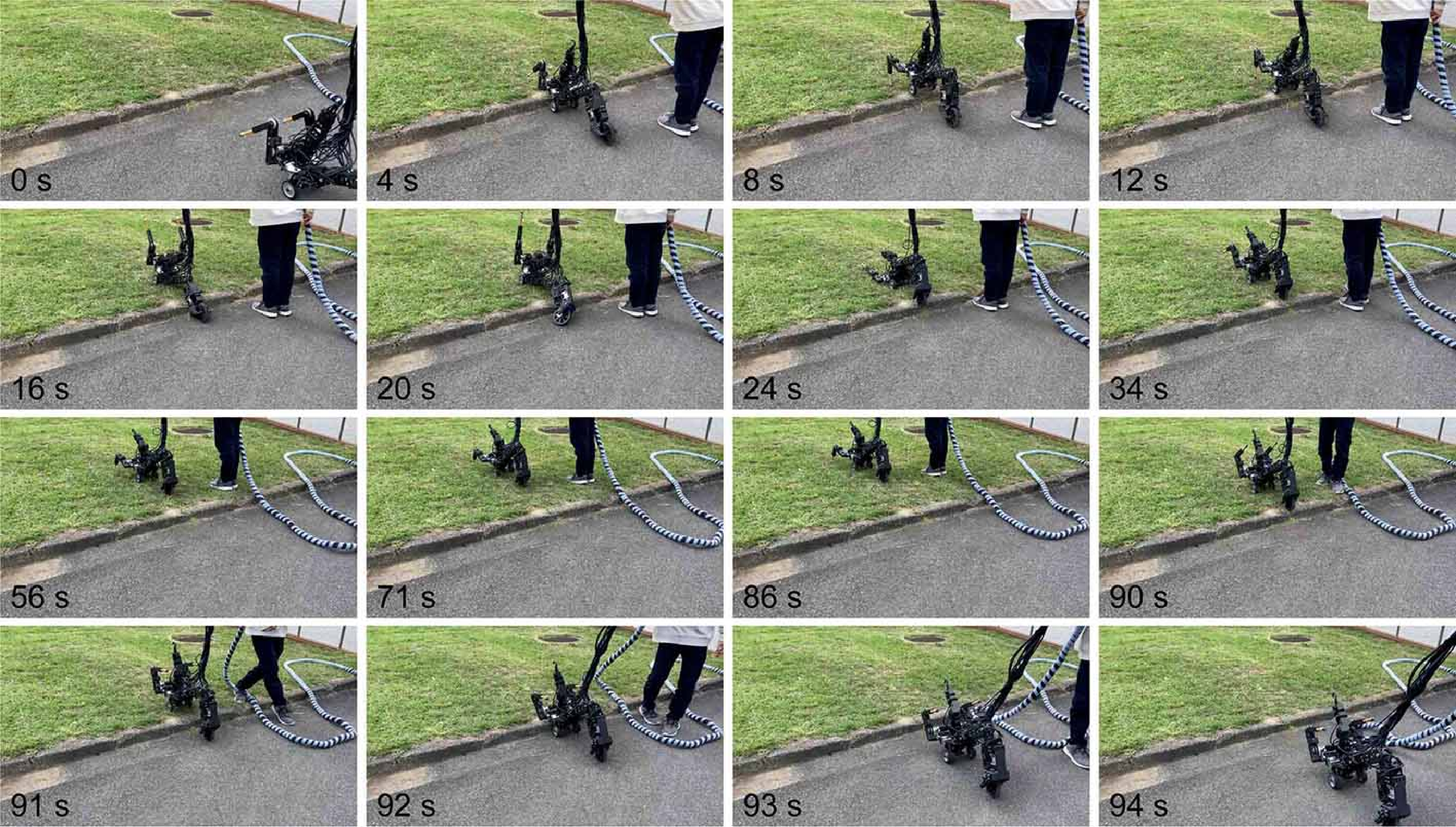}
	\end{center}
	\caption{Exp. 7: Outdoor locomotion.}\label{fig:exp:outdoor}
\end{figure*}
\section{Experiments} \label{sec:4}

We conducted the following seven experiments to verify \armss hybrid locomotion and loco-manipulation capabilities.
\begin{description}
	\item[Exp. 1:] \ \ Joint control
	\item[Exp. 2:] \ \ Wheeled-legged locomotion
	\item[Exp. 3:] \ \ Wheeled locomotion and grasping
	\item[Exp. 4:] \ \ Slope locomotion
	\item[Exp. 5:] \ \ Block terrain locomotion
	\item[Exp. 6:] \ \ Carrying a bag
	\item[Exp. 7:] \ \ Outdoor locomotion
\end{description}
%

The joint control experiment verified the trajectory modification and proxy-based sliding mode control with the angle, angular velocity, angular acceleration, and torque limits.
The verification used the joint $\theta_{32}$ of Arm~3, as shown in Fig.~\ref{fig:joint:exp}(a).
For this verification, the minimum joint limit was set to -1.75 rad, and the torque saturation was set to 60 \% of the maximum and minimum values.
The non-smooth command angle $\theta^\mathrm{cmd}_{32}$ was applied and yielded the impulsive velocity command $\dot{\theta}^\mathrm{cmd}_{32}$ and acceleration command $\ddot{\theta}^\mathrm{cmd}_{32}$, which exceeded the angle, angular velocity, and angular acceleration limits.
The command angle $\theta_{32}^\mathrm{cmd}$ was modified by the trajectory modification to the reference angle $\theta_{32}^\mathrm{ref}$, and the angle response $\theta_{32}$ tracked the reference angle with the proxy-based sliding mode controller \eqref{eq:PSMC}.
The reference angle $\theta_{32}^\mathrm{ref}$ was adjusted to satisfy the angle limits from 1 to 6 s, and the reference angular velocity and acceleration: $\dot{\theta}_{32}^\mathrm{ref}$ and $\ddot{\theta}_{32}^\mathrm{ref}$ also satisfied their limits at 1, 6, and 10 s, as shown in Fig.~\ref{fig:joint:exp}(b).
In 6--10 s, the arm contacted with the ground, as shown in Fig.~\ref{fig:joint:exp}(a), and did not reach the reference angle owing to the saturated torque $\tau_{32}$.
The angle response $\theta_{32}$ gradually returned to the reference angle $\theta^\mathrm{ref}_{32}$ after the saturation.
From 15 to 25 s, the angle response tracked the sine command precisely while the command did not exceed the limits.
%

In the wheeled-legged locomotion experiment, \arms moved forward with wheeled-legged walking, turned right from 6 to 21 s, and moved forward with wheeled locomotion from 22 s, as shown in Fig.~\ref{fig:exp:WheeledLeggedLocomotion}.
The steering joint $\theta_{13}$ for the active wheel and the walking of Arms 2 and 3 allowed turning, and \arms seamlessly changed the wheeled-legged locomotion to the wheeled locomotion with the active and passive wheels.
%

The wheeled locomotion and grasping experiment verified \armss loco-manipulation capability, as shown in Fig.~\ref{fig:exp:grasping}.
Arms~2, 3, and 4 grasped an orange plastic bucket given by an experimenter from 0 to 7 s.
Subsequently, \arms moved forward while grasping the bucket from 8 s.
Arms~2 and 3 were used as general-purpose arms for walking and grasping in Exps.~2 and 3, respectively.
%

\armss wheeled-legged locomotion was verified on a slope also.
Fig.~\ref{fig:exp:slope} shows the result.
\arms moved up the slope via the wheeled-legged locomotion from 0 to 56 s and moved down the slope via the wheeled locomotion from 57 s.
%

\armss legged locomotion capability enabled it to cross a deformable terrain, such as block terrain.
As shown in Fig.~\ref{fig:exp:block}, \arms moved forward over the many blocks by pushing against and partially lifting the torso using Arms~2 and 3.
It would be difficult to cross the block terrain with wheeled locomotion because the wheel can slip easily on blocks, and \armss hybrid wheeled-legged locomotion enhanced the locomotion capability on uneven terrain.
%

Fig.~\ref{fig:exp:carry} shows the result of the carrying a bag experiment.
\arms raised its torso with the three arms from 0 to 7 s, picked up the bag by using Arm~4 from 8 to 15 s, and lowered the torso from 16 to 22 s while holding the bag.
Subsequently, \arms carried the bag forward by wheeled-legged walking.
%

Lastly, \armss locomotion capability was verified in outdoor environment, as shown in Fig.~\ref{fig:exp:outdoor}.
\arms moved toward a step via the wheeled locomotion from 0 s to 4 s and stepped over the 80-mm step by lifting its torso and pulling Arm~1 from 5 to 35 s.
Then, \arms climbed a sloped lawn via the wheeled-legged locomotion.
Subsequently, it returned across the sloped lawn and the step via the wheeled locomotion.

\section{Conclusion} \label{sec:6}
This article proposed a mobile quad-arm robot: \arms capable of hybrid locomotion and loco-manipulation and provided the mechanical design, control design, and experiments of ARMS.
The experiments verified that \arms has the following capabilities: wheeled-legged tripedal locomotion; wheeled locomotion; locomotion on slope, block, and outdoor terrain; general-purpose capability of Arms~2 and 3; and grasping and carrying capabilities.
%

Compared with existing humanoid, quadrupedal, hexapedal, and hybrid locomotion robots, \arms is a better choice for scenarios combining fast locomotion on a flat terrain, stable locomotion on a flat or non-flat terrain, and object manipulation and requiring less actuators to minimize the cost and weight of the robot.
However, \armss manipulation capability is inferior to that of the humanoid and quadrupedal hybrid locomotion robots owing to its lower redundancy of the DOF.
Moreover, \arms has less stable locomotion than the quadrupedal and hexapedal robots because it has fewer legs.
Overall, compared with existing robots, \arms has the versatility with the less DOFs but is not specialized and is less redundant.
%

The developed robot has a limitation that needs to be resolved in the future research.
Although \arms requires servo amplifiers and a PC, the electrical system and the PC are not mounted on the robot.
This leads to a limited movable range of \arms owing to the motor cables, which are suspended from above, as shown in the experimental results.


%
\vspace{-2em}
\begin{IEEEbiography}[{\includegraphics[width=1in,height=1.25in,clip,keepaspectratio]{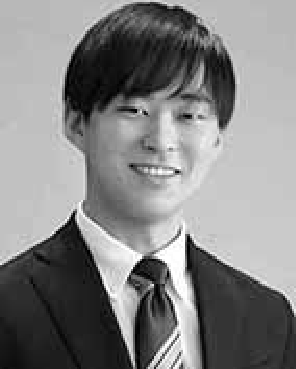}}]{Hisayoshi~Muramatsu}
received the B.E. degree in system design engineering and the M.E. and Ph.D. degrees in integrated design engineering from Keio University, Yokohama, Japan, in 2016, 2017, and 2020, respectively.
From 2019 to 2020, he was a Research Fellow with the Japan Society for the Promotion of Science.
He is currently an Assistant Professor with the Mechanical Engineering Program, Hiroshima University, Higashihiroshima, Japan.
His research interests include motion control, robotics, and physical human-robot interaction.
\end{IEEEbiography}
\vspace{-2.5em}
\begin{IEEEbiography}[{\includegraphics[width=1in,height=1.25in,clip,keepaspectratio]{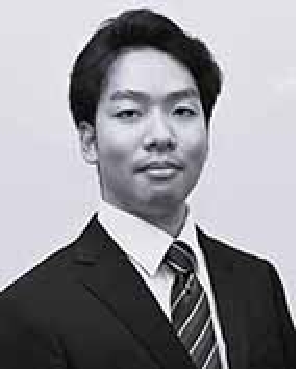}}]{Keigo~Kitagawa}
received the B.E. degree in mechanical engineering from Hiroshima University, Higashihiroshima, Japan, in 2022, where he is currently pursuing the master's degree in mechanical engineering with the Graduate School of Advanced Science and Engineering. His research interests include robotics and mechanical design.
\end{IEEEbiography}
\vspace{-2.5em}
\begin{IEEEbiography}[{\includegraphics[width=1in,height=1.25in,clip,keepaspectratio]{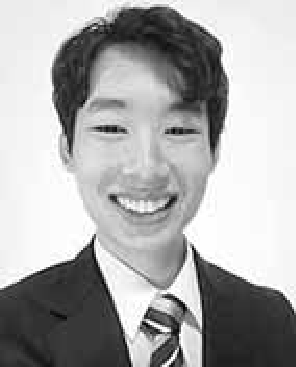}}]{Jun~Watanabe}
received the B.E. degree in mechanical engineering from Hiroshima University, Higashihiroshima, Japan, in 2023, where he is currently pursuing the master's degree in mechanical engineering with the Graduate School of Advanced Science and Engineering. His research interests include locomotion control of a mobile robot.
\end{IEEEbiography}
\vspace{-2.5em}
\begin{IEEEbiography}[{\includegraphics[width=1in,height=1.25in,clip,keepaspectratio]{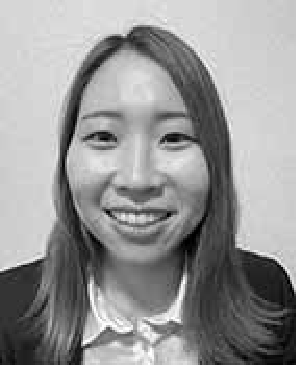}}]{Yuika~Yoshimoto}
is currently pursuing the B.E. degree in mechanical engineering at Hiroshima University, Higashihiroshima, Japan. Her research interests include robotics and mechanical design.
\end{IEEEbiography}
\vspace{-2.5em}
\begin{IEEEbiography}[{\includegraphics[width=1in,height=1.25in,clip,keepaspectratio]{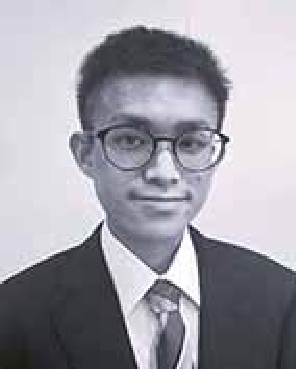}}]{Ryohei~Hisashiki}
received the B.E. degree in mechanical engineering from Hiroshima University, Higashihiroshima, Japan, in 2023, where he is currently pursuing the master's degree in mechanical engineering with the Graduate School of Advanced Science and Engineering. His research interests include mechanical design.
\end{IEEEbiography}

\vfill
\end{document}